\documentclass{article}

\usepackage[preprint]{neurips_2025}

\usepackage[utf8]{inputenc} 
\usepackage[T1]{fontenc}    
\usepackage{hyperref}       
\usepackage{url}            
\usepackage{booktabs}       
\usepackage{amsfonts}       
\usepackage{nicefrac}       
\usepackage{microtype}      
\usepackage{xcolor}         
\usepackage{comment}

\usepackage{amsfonts}       
\usepackage{siunitx}        
\usepackage{algorithm}      
\usepackage{hyperref}
\usepackage{url}
\usepackage{graphicx}
\usepackage[utf8]{inputenc} 
\usepackage[T1]{fontenc}    
\usepackage{hyperref}       
\usepackage{url}            
\usepackage{booktabs}       
\usepackage{amsfonts}       
\usepackage{nicefrac}       
\usepackage{microtype}      
\usepackage[title]{appendix}
\usepackage{marginnote}
\usepackage{amsmath} 
\usepackage{graphicx} 
\usepackage{algorithmic}  
\usepackage[algo2e, ruled]{algorithm2e} 
\usepackage[]{placeins} 
\usepackage{enumitem}
\usepackage{forest}
\usepackage[labelfont=bf]{caption}
\usepackage{listings}
\usepackage{cleveref}
\usepackage{mdframed}
\usepackage{bm}

\definecolor{mycolor}{RGB}{64, 111, 165}
\definecolor{mycolor2}{RGB}{188, 188, 69}
\definecolor{observed}{RGB}{239, 133, 54}
\definecolor{unknown}{RGB}{54, 160, 239}

\hypersetup{
  colorlinks,
  linkcolor={mycolor},
  citecolor={mycolor},
  urlcolor={mycolor}
}

\usepackage{tcolorbox}
\newtcolorbox{mybox}[1]{colback=unknown!5!white,colframe=unknown!75!black,fonttitle=\bfseries,title=#1}

\usepackage{tikz}
\usepackage{tikzscale}
\usepackage{wrapfig}

\usetikzlibrary{cd}
\tikzcdset{
  arrow style=tikz,
  diagrams={>={Straight Barb}}
}

\usetikzlibrary{patterns,positioning,arrows,shapes,shapes.geometric,shadows}
\usetikzlibrary{decorations,decorations.pathmorphing,intersections,backgrounds}

\usepackage{pgfplots}

\DeclareMathOperator*{\argmax}{arg\,max}

\newcommand{\children}[1]{\mathrm{children}{#1}}

\newcommand{\Dir}{\mathrm{Dir}}

\newcommand{\vc}{\pmb{c}}
\newcommand{\vv}{\pmb{v}}

\newcommand{\methodname}{ULTS}
\newcommand{\methodlongname}{Uncertainty-guided Likelihood-Tree Search}

\def\gA{{\mathcal{A}}}

\def\gS{{\mathcal{S}}}

\title{Uncertainty-Guided Likelihood Tree Search}

 \author{\textbf{Julia Grosse} \thanks{julia.grosse@uni-tuebingen.de}\\
 University of Tübingen, Tübingen AI Center
  \And
  \textbf{Ruotian Wu}\\
 University of Waterloo
  \And
  \textbf{Ahmad Rashid}\\
 Vector Institute, University of Waterloo
    \And
  \textbf{Cheng Zhang}\\
 GenAI at Meta
  \And
\textbf{Philipp Hennig}\\
 University of Tübingen, Tübingen AI Center
  \And
 \textbf{Pascal Poupart}\\
  Vector Institute, University of Waterloo
   \And
  \textbf{Agustinus Kristiadi}\\
 Vector Institute
 }

\begin{document}

\maketitle

\begin{abstract}
Tree search is a fundamental tool for planning, as many sequential decision-making problems can be framed as searching over tree-structured spaces. 
We propose an uncertainty-guided tree search algorithm for settings where the reward function is a log-likelihood function of the paths. 
Due to the combinatorial explosion of the tree size, the set of paths for which one can obtain rewards is sparse, particularly when the likelihood is obtained through expensive evaluations, such as by querying a large language model. 
We address this challenge by deriving a probabilistic search heuristic based on regularity assumptions for the likelihood. 
Unlike existing tree search methods, the proposed method can perform backtracking and trade-off exploration with exploitation, and yet does not require expensive roll-outs, or sophisticated Bayesian inference.
Through extensive on-model and off-model experiments on timely, large-scale practical applications, we demonstrate that our method identifies paths with high likelihood while requiring fewer costly evaluations.

\end{abstract}

\section{Introduction}
\label{sec:intro}

Searching over tree-structured spaces is ubiquitous, for example in industrial scheduling \citep{lubosch2018industrial}, games \citep{silver2017mastering}, chemistry \citep{corey1989}, and robotics \citep{guruji2016time}.
Essentially, the goal is to find a path in the search tree that maximizes some notion of reward.
In general, this problem is intractable: 
The number of possible paths is exponential, especially when the branching factor and the depth are large. 
To make matters worse, the evaluation of each node's reward is often expensive.
Thus, searching over such a tree often exceeds the computational budget required to examine them all.

This inevitably leads to \textit{computational uncertainty} \citep{hennig2022probabilistic}: 
an uncertainty that \textit{could be} fully resolved if enough compute was available to examine all paths, but in practice is present due to the limited resources.
Standard search algorithms for large-scale search trees, such as beam search \citep{koehn2003statistical}, completely ignore this uncertainty. 
Meanwhile, uncertainty-aware algorithms such as Monte Carlo tree search \citep[MCTS,][]{kocsis2006mcts} and its Bayesian counterparts \citep{hennig2010coherent,tesauro2012bayesian,mern2021bayesian} require repeated (costly) evaluation of nodes, additional expensive models, or complicated approximate inference.

To maximize efficiency and cost-effectiveness, we are thus interested in an online tree-search algorithm that can exploit computational uncertainty without rollouts, expensive models, and complicated Bayesian inference.
We posit that even a \emph{simple} way of quantifying the aforementioned uncertainty can enhance exploration within the search tree; ultimately reducing the number of node expansions while maintaining the same or achieving a better reward. 
To this end, we incorporate computational uncertainty into the search process to guide it in a non-myopic fashion---i.e., accounting for our \emph{belief} about the values of future nodes, taking inspiration from how humans might plan \citep{baumeister2016pragmatic}---and in a data-efficient manner, akin to Bayesian optimization \citep{kushner1964bayesopt,movckus1975bayesopt}.

Bayesian optimization is recognized for its efficiency not merely because it quantifies uncertainty, but also because it exploits the structural characteristics of the problem.
E.g., in continuous optimization, prior knowledge, such as the smoothness of a function, is often available through Gaussian processes \citep{rasmussen2003gaussian} or neural networks \citep{hernandez2017parallelTS,kristiadi2023promises}.
In our setting, we have the structural assumption that the immediate reward at each node of the search tree is a component of a Categorial distribution, i.e.~normalized in \([0, 1]\).
Intuitively, the reward of each node is the \emph{likelihood} of the node being in a given root-to-leaf path in the tree, and the goal is to find a path with the highest total likelihood (cumulative reward).
This setting is of great interest since it occurs in problems such as autoregressive generation processes like in large language models.

In our setting, the \emph{a priori} characteristic property we aim to exploit is the \textit{concentration strength}: whether the Categorical distributions are all highly concentrated at a few realistic options, or if some of them are nearly uniform, making any individual options drastically less likely to be the optimal one.
Intuitively, one would expect this to have a strong influence on the number of paths that need to be considered.
For instance, when the distribution is concentrated, it is less likely under our belief that other paths will overtake later on and one can be more greedy.
Meanwhile, when the distribution has high entropy, the uncertainty of our belief about ``which next child node is best'' is higher and thereby requires more exploration and computational budget.

In this work, we propose a probabilistic framework that captures this aspect of the search space, which can help to decide which paths should be pursued and which can be ignored based on the posterior belief.
We do so by using a non-myopic acquisition function based on the samples of such a belief, which can be seen as a generalization of Thompson sampling \citep{thompson1933sampling}.
Crucially, these samples are \emph{cheap} to obtain, unlike other empirical sample-based tree-search methods like MCTS.

Ultimately, building upon elements from beam search, MCTS, and Bayesian tree-search methods, we propose an uncertainty-aware heuristic search called \emph{\textbf{\methodlongname{} (\methodname{})}}.
This method only adds a small runtime overhead relative to the likelihood/reward evaluation in scenarios where such a likelihood is obtained through a costly model evaluation since neither rollouts, additional expensive models, nor elaborate Bayesian techniques are present in ULTS.
Nevertheless, in experiments involving large language models, we show how ULTS can generally achieve a favorable cost-performance tradeoff compared to standard baselines.
In sum:
\begin{enumerate}[label=\roman*., itemsep=0em, topsep=0em]
  \item We propose \methodname{}: a probabilistic heuristic search for non-myopic sequential decision-making with step-wise likelihood rewards. 
  We leverage samples from a prior over the reward and show how to easily sample from the implied posteriors over future nodes' values. 
  These samples are useful to efficiently expand the search tree.
  \item We demonstrate the efficiency and extensibility of \methodname{} in on- and off-model experiments.
  In the latter, we assume the likelihood of the paths is obtained from querying large language models.
    \item We open-source a Python implementation of ULTS, which is compatible with \texttt{transformers} \citep{wolf-etal-2020-transformers} for large language model applications that is available at \url{https://github.com/JuliaGrosse/ults}.
\end{enumerate}

\section{Setting}
\label{sec:background}

Formally, our setting can be viewed as a specific instance of a Markov decision process (MDP), characterized by a tree-structured state space with deterministic transitions and a constraint on the also deterministic reward function. Let 
\( (\mathcal{S}, \mathcal{A}, P, R, \gamma) \)
be a MDP,\footnote{We assume a discount factor $\gamma=1$ for simplicity, but the discussion can easily be extended to include other values as well.} where \(\gA = \{ a_1, \dots, a_B \}\) is a finite set of actions and \(\gS = \bigcup_{d=0}^D \gA^d \) the set of states containing all sequences of actions up to length $D$, the function \( P(s' \mid s, a) \) given by
\begin{align*}
P(s' \mid s, a) =
\begin{cases}
1 & \text{if $s' = \text{concat}(s,a)$}\\
0 & \text{otherwise,}
\end{cases}
\end{align*}
is the transition probability, and \( R(s, a) \) is the reward function, giving the immediate reward received for taking action \( a \) in state \( s \). 
We assume that at each non-terminal state $s_i$, there is a Categorical distribution $\vc_i : \mathcal{A} \rightarrow [0,1]$ available that models the \textit{myopic} probability for the next action to be the best choice. 
We have the constraint $\vc_i(a) \geq 0$ and $\sum_{a \in \gA} \vc_i(a) = 1$. 
In order to obtain the usual additive structure of the rewards, we optimize in log space, i.e. $R(s_i,a) = \log \vc_i(a)$.
Thus, the \(R\) in this setting can be interpreted as (log) likelihood, and we call the search space a \emph{likelihood tree}.

The goal is to find a path $s_0 \to s_D := (s_0, ..., s_D)$ from the root node $s_0$ to a terminal state $s_D \in \gA^D$ that maximizes the cumulative rewards $c_{s_0 \to s_D} := \exp \sum_{d=1}^D R(s_d,a_d)$, i.e. the (log) \emph{likelihood} of the path.
Specifically, we assume the setting where evaluating \(R(s, a)\) is \emph{expensive}, and the depth \(D\) and branching factor \(B\) of the tree are \emph{large}.
Lastly, we define recursively the optimal value $v_i$ of a node $s_i$ as the sum of the logarithm of the transition probabilities $c_{s_0 \rightarrow s_i}$ from the root node $s_0$ to $s_i$ and a remaining term which we refer to as $\log \Delta_i$, i.e., we have $v_{s_i} = \log c_{s_0 \rightarrow s_i} + \log \Delta_i$. Intuitively, the term $\Delta_i$, quantifies the likelihood that we get in the remaining steps from $s_i$ to a leaf node when we take all remaining decisions optimally.
It can be defined by the following recurrence relation:
\begin{align}
    \Delta_i = \begin{cases}
                    1                                                                 & \text{if $s_i$ is a leaf,} \\
                    \underset{s_j \in \children{s_i}}{\max} \{\vc_{i}(a_j) \cdot \Delta_j\} & \text{otherwise.}          \
                  \end{cases}
\end{align}

\textbf{Example: Decoding Large Language Models} \quad Here, the set of actions \(\gA\) is a vocabulary consisting of (natural) language tokens.
The root of the search tree is given by the context, e.g., a question in a question-answering system.
The reward function \(R\) is determined by a forward pass through a large language model (LLM), i.e. a large-scale neural network that utilizes the attention mechanism \citep{vaswani2017attention} to predict the probability of the next token appearing in a sequence. 
This process is done recursively until termination; either when a specified depth \(D\) has been reached or when a special token ``\(\langle \text{EOS} \rangle\)'' is selected. Considering the vocabulary size \(B\) ranges from around \(32\)k to \(256\)k \citep{radford2019language,chowdhery2023palm} and the reward evaluation amounts to a costly forward pass, even when the sequence length \(D\) is small, computing the optimal sequence using standard tree-search algorithms like A$^{*}$ or Monte Carlo tree is search infeasible.
Thus, in practice, one is limited to using heuristic algorithms that are close to greedy, e.g.\ beam search and its variants \citep{vijayakumar2016diverse, meister2021determinantal, freitag2017beam}. 

Beyond LLM decoding, several MDPs fit within our problem formulation, where stepwise likelihood serves as a reward. In general, ULTS is applicable to any autoregressive model, not just in the language domain, but for example also in chemistry tasks \citep{christofidellis2023unifying, gao2024generative}. Further specific example applications include optimizing process parameters for machines (such as paper drying machines in \citep{chen2025reinforcement}), feature selection \citep{luo2019autocross}, various neural combinatorial optimization problems by decoding pointer networks \citep{vinyals2015pointer}, structure learning tasks \citep{kok2005learning}, and finding Maximum A Posteriori sequences as in \citep{pellom2002efficient}. 

\section{Method}
\label{sec:method}

Here, we introduce \textbf{\methodlongname{} (\methodname{})} which places a prior belief over the log-likelihood reward function and computes the implied posterior samples over nodes' values.
We introduce and discuss the modeling assumptions in \cref{model-likelihood-rewards} and show how to derive approximate beliefs over the optimal values in the search tree in \cref{optimal values}.
The samples of the posterior beliefs are then used to make decisions about which subtree to expand next and when to stop the search (\cref{acquisition function}).

\subsection{Probabilistic model}

\subsubsection{Prior beliefs over immediate rewards}
\label{model-likelihood-rewards}

A straightforward belief one can consider is the Dirichlet distribution.
For tractability, we assume that the probabilities are iid.~draws from a symmetric Dirichlet distribution with parameter $\alpha > 0$, i.e., \(p(\vc_s) = \Dir(\alpha)\).
Thus, $\alpha$ controls how concentrated the sampled probability vectors are.
For small $\alpha$, one would typically strongly favor a few children, whereas for large $\alpha$ the discrete distribution would closely resemble a uniform distribution over the children. 
The symmetry of the prior implies in our context that we do not have a preference for particular children \textit{a priori}, but it is also possible to replace the symmetric prior with an asymmetric one with no additional costs.

In settings where the Categorical distributions are predicted by a trained sequence model $f$, an alternative option is to define an empirical prior over \(\vc_i\) based on predictions $\vc_i = f(s_i)$ on a subset of the training/validation data. 
Let $\smash{\{s_{0,n}\}_{n=1}^{N}}$ be $N$ samples of root nodes from this subset.
We can obtain a set \(\{s_{0,n}, s_{1,n}, \dots, s_{D,n} \}_{n=1}^N\) of \(D\)-step completions of \(s_{0,n}\)'s with $f$ (e.g.\ through a greedy search).
We can then collect samples of the Categorical distributions from this generation process: \(\mathcal{C} = \{ \vc_{0,n}, \dots, \vc_{D-1,n} \}_{n=1}^N\).
Then, instead of sampling from \(\Dir(\alpha)\), we can sample from \(p(\vc_i) = \mathrm{Unif}(\mathcal{C})\).
This prior is more flexible than the Dirichlet prior since no symmetry nor unimodality assumptions are made at the cost of more overhead.
In any case, all these priors are \emph{precomputed}, so they incur a fixed \(\mathcal{O}(1)\) cost.

\subsubsection{Prior beliefs over values}
\label{optimal values}

Having picked a prior for the categorical distributions, we can derive the implied priors over the optimal values $v_{s_i}$ for each node \(s_i\) in the tree \citep{hennig2010coherent}. As introduced in Section \ref{sec:background}, we write the optimal value of a node as $v_{s_i} = \log c_{s_0 \rightarrow s_i} + \log \Delta_i$.
Due to the iid.~assumption above (Section \ref{model-likelihood-rewards}), we have the joint distribution $p(c_{s_0 \rightarrow s_i}, \Delta_i) = p(c_{s_0 \rightarrow s_i})\, p(\Delta_i)$. While all these quantities are not analytically available, we shall show that sampling from the \emph{posterior} belief \(p(v_{s_i} \mid c_{s_0 \to s_i})\) is easy if we are able to sample from \(p(\Delta_i)\). Let us, therefore, derive an approximate sampling scheme for $p(\Delta_i)$.
It follows \cite{hennig2010coherent, grosse2021probabilistic} who used Gaussian priors.
We recursively approximate the prior distribution of the $\Delta_i$'s at level \(l\) with Beta distributions $\text{Beta}_{l}(\Delta_i)$.
In a bottom-up approach, we generate \(N\) samples $\{\max_j \vc_{n}(a_j) \mid \vc_n \sim p(\vc) \}_{n=1}^N$ for a $\Delta_i$ starting at level $l=D-1$.
Using these samples, we empirically fit the parameters of the Beta distribution $\mathcal{B}_{D-1}(\Delta_i)$ via maximum likelihood.
The distributions of the $\Delta_i$ are the same for all nodes on the same level due to the iid.\ assumption, so this has to be done only once.
Note that we need this approximation since the distribution of the maximum \(\max_j \vc_{i}(a_j)\) has no known analytic solution.
We then continue by recursively sampling sets of the form (one per level)
$\{\max_j \vc_{n}(a_j)\cdot \Delta_j \mid \vc_n \sim p(\vc), \Delta_j \sim \mathcal{B}_{l+1}(\Delta) \}_{n=1}^N$
for a $\Delta_i$ of the level $l$ and using it to fit the parameters of $\mathcal{B}_{l}(\Delta_i)$.
The time complexity is $\mathcal{O}(D B N)$ for computing the approximations, i.e.,\ it is linear in the depth and width of the tree.
We emphasize that they can be pre-computed before the search and reused across different decoding runs. 
Alg.~\ref{alg:beta} shows the pseudocode. 

\begin{algorithm}
    \caption{Prior belief over \(\Delta\)}
    \label{alg:beta}
    \begin{algorithmic}[1]
        \STATE \textbf{Input:} Depth of the tree $D$, branching size of tree $B$, prior \(p(\vc)\) over categorical rewards, number of samples used for the approximation $N$
        \STATE \textbf{Output:} Table of parameters $\text{params}_{l}$ for Beta distributions $\mathcal{B}_{l}$ for each level $l$ of the tree
        
        \FOR{$l \gets D-1$ to $1$}
            \FOR{$n \gets 1$ to $N$}
                \STATE Sample $\vc_n \sim p(\vc)$
                
                \FOR{$j \gets 1$ to $B$}
                    \IF{$l = D-1$}
                        \STATE $\Delta_{nj} \gets 1$
                    \ELSE
                        \STATE Sample $\Delta_{nj} \sim \mathcal{B}_{l+1} (\Delta \mid \text{params}_{l+1})$
                    \ENDIF
                \ENDFOR
                
                \STATE $\Delta_n \gets \max\limits_{j=1,\dots,B} (\vc_{nj} \cdot \Delta_{nj})$
            \ENDFOR
            
            \STATE $\text{params}_{l} \gets \texttt{beta-MLE}(\{\Delta_{n}\}_{n=1}^{N})$
        \ENDFOR
    \end{algorithmic}
\end{algorithm}

\subsubsection{Posterior beliefs on frontier nodes}
\label{subsec:posterior}

Notice that whenever a new node $s_i$ is added to the search tree, the likelihood associated with the path from the root to that node $c_{s_0 \rightarrow s_i}$ is fully observed---its distribution is simply a Dirac delta.
This means, the joint distribution becomes \(p(\vc_{s_0 \to s_i}, \Delta_i) = \delta(c_{s_0 \to s_i}) \, p(\Delta_i)\).
Therefore, to sample \(v_{s_i}\) given that we have observed \(c_{s_{0} \to s_i}\)---i.e., sampling from the posterior $p(v_{s_i} \mid c_{s_0 \rightarrow s_i})$---it is sufficient to sample from $p(\Delta_i)$ and then simply scale all samples by $c_{s_0 \rightarrow s_i}$. During the search, these samples are backed up the tree and leveraged to make decisions in exploring the tree.
Notice that these samples are cheap to get, unlike empirical samples based on simulations/rollouts.

\subsection{Algorithm}
\label{acquisition function}

At each iteration, we use the posterior samples by following the steps of Monte Carlo tree search, but without the expensive rollout step.
Pseudocode in \cref{apd:pseudocode}.

\textbf{Step 1: Selection} \quad Starting from the root node \(s_0\), we recursively pick a child
based on an acquisition function until an unexpanded node is selected. From a decision-theoretic perspective, this is done by choosing a utility function that encodes our preferences about the outcome (e.g. high likelihood) and then integrating out the unknown variables influencing this outcome. 
Let \(\vv_i\) contain the (unknown) optimal values of the descendants of \(s_i\) and let \(u(s_c, \vv_i)\) be a utility function. The idea is to select the child node $s_c$ that maximizes the expected utility:

\begin{equation} \label{eq:acqf_integral}
  \textstyle \int u(s_c, \vv_i) \, p(\vv_i \mid c_{v_{s_0 \to s_i}}) \, d\vv_i ,
\end{equation}

For ULTS, we use an utility function that encodes our preference for finding a descendant with high optimal value: \(u(s_c, \vv_i):= \mathbb{I} [ v_{\hat{s}_c} = \max_{s_j \in \mathrm{children}(s_i)} v_{\hat{s}_j} ]\). Different realizations of \(\hat{s}_c\) and \(\hat{s}_j\) can be used:
The most straightforward is to use the children's optimal values themselves.
This corresponds to simply setting \(\hat{s}_j = s_j\) and \(\hat{s}_c = s_c\).

Another strategy is to use the optimal values of each child's best descendant---intuitively, it ignores other paths in the tree that might have higher values, leading to a more pessimistic belief.
This can be beneficial to bias the search towards more exploration (for further discussion, see \cref{apd:posterior_acqf}).
In this case, \(\hat{s}_j\) is defined as (similarly for \(\hat{s}_c\)):
\begin{equation}\label{eq:acqf_recurrence}
  \scalebox{0.85}{
    $\hat{s}_j = \begin{cases}
        s_j                                                         & \text{if } s_j \text{ is a leaf,} \\
        \underset{s_c \in \mathrm{children}(s_j)}{\argmax} a_j(s_c) & \text{otherwise}.
      \end{cases}$
  }
\end{equation}
Note that they have the same costs since beliefs over the optimal values for both strategies are readily available due to the \textsc{backup} step below, i.e., we do not actually perform the recursion \eqref{eq:acqf_recurrence} in this step.

The acquisition function is then derived as a sample-based 
approximation to the expectation in \eqref{eq:acqf_integral}:
\begin{equation}\label{eq:acqf}
  \scalebox{0.85}{$a_i(s_c) = \frac{1}{N} {\displaystyle \sum_{n=1}^N} \mathbb{I}\bigl[v_{\hat{s}_c, n} = \underset{s_j \in \mathrm{children}(s_i)}{\max} v_{\hat{s}_j, n}  \bigr] $}.
\end{equation}
Note that we can replace the above utility function with \emph{any} other utility function and then use the posterior samples to derive the acquisition function \citep{wilson2018maximizing}. For example, we can use a utility function that encodes a preference for sequences with fewer repetition, see \cref{apd:penalty}.

\begin{figure}
\includegraphics[]{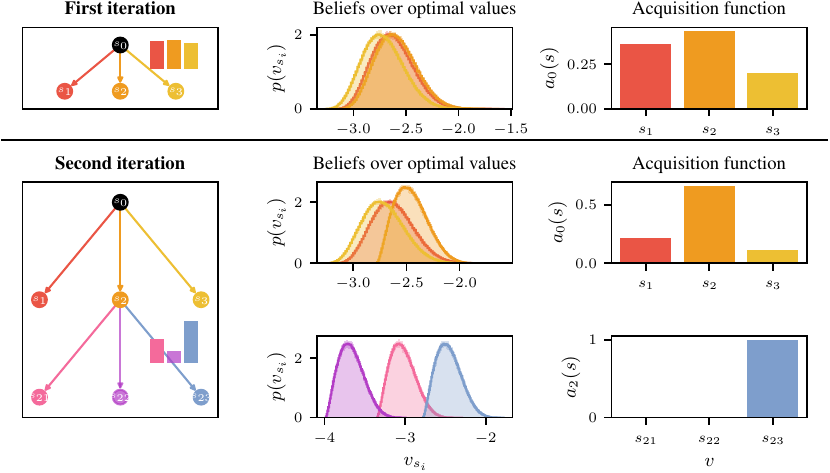}

\vspace{-0.5em}

\caption{
An example of two iterations with ULTS. The upper row show the observed categorical distirbution (left), the implied posterior over the optimal values in log space (center) and the resulting acquisition function over the children in the first level of the tree (right).  The lower two rows show the corresponding quantities for the first and second level of the tree after the second iteration.
}
\label{fig:ToyExample}
\end{figure}

\textbf{Step 2: Expansion} \quad Given an unexpanded node $s_i$, we evaluate its children's rewards.
These newly obtained rewards are new observations that can be combined with the prior samples to obtain \(N\) posterior samples $\{v_{s_c,n}\}_{n=1}^N$ of each child $s_c$.
In turn, they will affect the acquisition function in the next iteration.

This step is often the most expensive part of any tree-search algorithm since we assume costly evaluation of the rewards.
Fewer node expansions is thus preferable.

\textbf{Step 3: Backup} \quad We recursively propagate the newly obtained posterior samples $\{ v_{s_c,n} \}_{n=1}^N$ back up the tree until the root via the path selected in the previous steps.
This is done to update the posterior samples contained in each node of the path.
They will then influence the selection process in the next iteration, updating the exploration-exploitation tradeoff.

Different update strategies can be used depending on the choice of the acquisition function \eqref{eq:acqf}.
When posterior samples of the children node \(s_c\) are used to compute \(a_i(s_c)\), then we propagate up the posterior samples of the newly expanded nodes as in when computing the prior by recursively taking their maximum, i.e. the $n$-th sample for the optimal value of parent node $s_p$ is given by the maximum over the $n$-th sample of the children's optimal values:
\begin{equation}
v_{s_p,n} = \max_{s_c \in \children(s_{p})}  v_{s_c,n}.
\end{equation}
When the posterior samples of the best descendant of \(s_c\) are used in \(a_i(s_c)\), we simply propagate up the posterior samples of the best child among the newly expanded nodes, without taking further maximums along the path:
\begin{equation}
v_{s_p,n} = v_{s_c*,n},\quad \text{where }s_c^* = \underset{s_c \in \mathrm{children}(s_p)}{\argmax} a_p(s_c).
\end{equation}
\Cref{fig:ToyExample} shows examples for the first two iterations of ULTS under the former acquisition function and backup scheme.
See \cref{apd:posterior_acqf} for the corresponding illustration for the latter.

\textbf{Step 4: Termination} \quad The posterior samples over the optimal values can not only be used for the selection of new nodes but also to monitor the progress of the optimization.
For instance, one can compute the following empirical probability
$\hat{\mathbb{P}}(c^* < v_{s_0}) = \frac{1}{N} \sum_{n=1}^N \mathbb{I}\left[c^* \leq  v_{s_0,n}\right]$,
which corresponds to the probability of the current best likelihood \(c^*\) among all leaves ULTS has visited so far is lower than the best value \(v_{s_0}\) at the root, according to our posterior belief.

Note that the computation of such a probability is done as in the acquisition function \(a(x)\), i.e., using the posterior samples of the root node $s_0$ or the posterior samples of the best descendant.
Then, one can decide to stop the search once this probability is below some confidence level $\varepsilon > 0$.

\subsection{Remarks}

\textbf{Practical considerations} \quad
In order to put a tractable upper bound on the runtime of ULTS, we introduce a hyperparameter $k_{\max}$ on the maximum number of nodes that can be expanded per level, similar to beam search.
Moreover, we stop the search as soon as a set of $k_{\max}$ leaves is attained or the termination probability exceeds below $1-\varepsilon$.
We shall see in \cref{sec:experiments} that even under these further constraints, we still obtain good results while being efficient.

\textbf{Limitations} \quad
In this work we focus on the uncertainty arising from the exponentially growing search space under limited compute resources. However, there might be other sources of uncertainty, e.g. uncertainty about the likelihood/reward function \(R\). In LLM decoding, there is evidence that higher likelihood is not always correlated with human preferences \citep{holtzman2019curious,wiher2022decoding, stahlberg2019nmt, eikema2020map, zhang2020trading}. Appendix \ref{apd:add_results} contains additional evaluations of ULTS in terms of other metrics than the likelihood, where there is indeed no perfect correlation. There is an ongoing line of research on how to counteract miscalibration in LLMs that is complementary to our work \cite{yang2023bayesian, kapoor2024large, onal2024gaussian, tonolini2024bayesian}, but combining and evaluating such approaches with ULTS goes beyond the scope of this paper. 
Moreover, the iid.\ prior assumption might not fully hold, but in our inference scheme it is essential to obtain tractability.
This is akin to how priors in Bayesian neural networks \citep{wilson2020bayesian} are usually chosen, which are just iid. Gaussians.

\section{Related Work}
\label{sec:related}

Probabilistic tree search methods have been explored for various applications.
\cite{hennig2010coherent,tesauro2012bayesian} introduced a method for game trees like Go, using a Gaussian process prior and expectation propagation \citep{minka2001family} to model the beliefs. 
\cite{grosse2021probabilistic} extended this to directed acyclic graphs. 
\cite{mern2021bayesian} also used Gaussian processes for general planning problems.
Unlike these methods, we focus on trees with a Categorical likelihood reward function. 
Additionally, the above methods rely on MCTS-based roll-outs, which do not scale efficiently. 
ULTS avoids costly rollout statistics by leveraging precomputed belief samples for decision-making.

Our approach employs a best-first search strategy, akin to A* \citep{hart1968astar}. An A*-like beam search algorithm, under the name of best-first beam search, has also been studied in \cite{meister2020best}. 
Unlike the non-probabilistic A$^*$ and best-first beam search, we approach optimization-on-tree problems via the lens of decision-making under uncertainty---putting a prior belief about the unknown, updating it based on the observations, and making decision based on the posterior belief.

MCTS is a staple uncertainty-aware search method and it has also been proposed for various LLM applications \citep{leblond2021machine, liu2023don, hao2023reasoning, feng2023alphazero, zhang2023planning, zhou2023language, delorenzo2024make}.
However, unlike ULTS, MCTS is non-probabilistic and generally costly since it requires rollouts.
Moreover, it is often used for cases where rewards are only observable at leaf nodes \citep[etc.]{zhang2023planning} or when external value estimators is available \citep[etc.]{feng2023alphazero}.
The latter introduces a substantial overhead in evaluating a value network, which is often another large-scale model.
ULTS, meanwhile, uses a simple model which introduces negligible overhead (see \cref{fig:runtime}).

Minimum Bayes risk methods \citep{kumar2004minimum, eikema2021sampling, bertsch2023itsmbrwaydown} also rely on the maximization of an expected utility function.
However, they do not employ probabilistic models and do not aim to make decisions in a sequential manner.
Rather, they generate a set of full-sequence samples and make a decision on which sequence in the set to be selected. 

\section{Experiments}
\label{sec:experiments}figures

In this section, we evaluate ULTS in both on-model and off-model problems.
In the former, we assume that the rewards are sampled from a Dirichlet distribution.
In the latter, we use black-box LLMs as reward functions. Unless specified otherwise, we use the acquisition function in \eqref{eq:acqf_recurrence} for the selection and backup steps.
See \cref{apd:add_results} for results with the other strategy and further dataset/setting.
All experiments are done on a single NVIDIA GeForce RTX 2080 Ti and NVIDIA A40 48GB GPUs for GPT-2 and Llama-2-7b, respectively. Errorbars reported in the plots and tables indicate the standard error of the mean as calculated with \texttt{scipy.stats.sem}. 

\subsection{On-model experiments}

We compare \methodname{} to beam search on artificially generated search problems from Dirichlet priors.
The trees have branching factor $B=8$ and depth $D=5$. 
Since these trees are so small we optimized the acquisition function in \eqref{eq:acqf} over the entire frontier, i.e.\ non-recursively with $k_\text{max} = \infty$ like standard A*.
The transition probabilities at each node are sampled from a Dirichlet prior with fixed $\alpha \in \{0.1, 0.2, 0.5, 0.8\}$.
The comparison is on-model, i.e., \methodname{} is run with the ground truth parameter of $\alpha$.
We repeat the experiment with different values for the confidence parameter \(\varepsilon\) of \methodname{} from $\{0.05, 0.1, 0.3, 0.5\}$.
Beam search is run with beam sizes ranging from \(1\) to \(7\). Since the $\Delta$-Terms of ULTS play the same role as the heuristic terms in A*, we also add A* with uninformative, optimistic heuristic (i.e. estimating the probability of the remaining sequence parts with 1) for an additional layer of ablation. 
The results in \cref{fig:synthetic} and \cref{fig:dirichletsamples} in Appendix \ref{apd:on_model} show that \methodname{} dominates across the entire range of hyper-parameters.
This suggests that knowledge about concentration strength helps reduce the number of search steps.

\begin{figure}
  \includegraphics[]{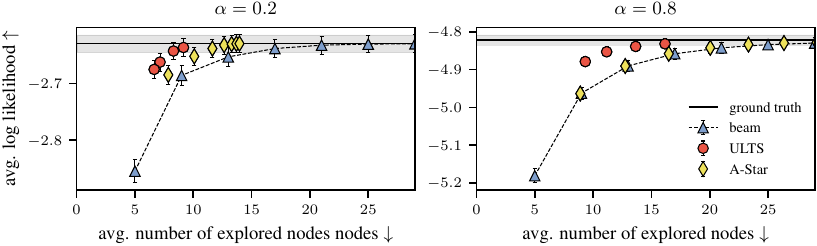}
  \caption{
    Comparison of ULTS with beam search and A* on synthetic rewards sampled from a Dirichlet distribution with $\alpha=0.2$ and $\alpha=0.8$.
    ULTS finds the optimum efficiently. Errorbars indicate one times the standard error of the mean. 
  }
  \label{fig:synthetic}
\end{figure}

\subsection{Off-model experiments with LLMs}

\textbf{Setup} \quad
We set ULTS' $k_\text{max} \in \{2,3,4,5,10,20\}$ and set $\varepsilon$ to a default value of $0.1$.
The error bars indicate \(\pm 1\) standard error of the mean based on all test sentences.
We compare ULTS against beam search, as well as other recent baselines commonly used for LLM decoding: multinomial beam search \citep[Multinomial BS;][]{kool2019stochastic},
contrastive search \citep{su2022contrastiveframeworkneuraltext} nucleus sampling \citep{holtzman2019curious}, best-of-$k$ sampling \citep{stiennon2020learning}, speculative decoding \citep{leviathan2023fastinferencetransformersspeculative}, and DoLA \citep{chuang2024doladecodingcontrastinglayers}.
For beam search, multinomial beam search, and best-of-k baselines, we evaluate beam sizes and numbers of samples \(k \in \{ 1, 2, 3, 4, 5, 10, 20 \}\), respectively.
We set any other hyperparameters of all baselines as suggested by the original papers or by the \texttt{transformers} library. 
We use Huggingface's implementation of the baselines (except for A* which is not part of Huggingface's repertoire). We use GPT-2 \citep{radford2019language} and Llama-2-7b \citep{touvron2023llama2} for text generation on articles from the Wikipedia \citep{see-etal-2017-get}, CNN Daily Mail \citep{hermann2015teaching}, and Reddit TL;DR \citep{volske2017tldr} datasets.\footnote{Additional results on machine translation tasks can be found in \cref{apd:rouge}.}
Since many of the text samples in the Wikipedia dataset end with e.g., references instead of full sentences, we filter for text samples with at least \(500\) tokens, resulting in a test set with \(332\) token sequences (out of a random subset of originally \(1000\) sequences).
We use \(200\) tokens as input and predict \(20\) tokens.
We do the same for the CNN Daily Mail dataset, where we end up with \(790\) token sequences.
We use a context length of \(300\) and generate \(60\) tokens.
We also include a summarization task, where the goal is to generate a \(40\)-token long summary of the input sequence.
For this, we use \(1000\) random samples from the TL;DR dataset with variable-length contexts. 
We run ULTS with both the Dirichlet prior with $\alpha=10^{-4}$ and the empirical prior. For more details regarding the choice of prior, see \cref{apd:prior}.

\begin{figure}[t]
  \includegraphics{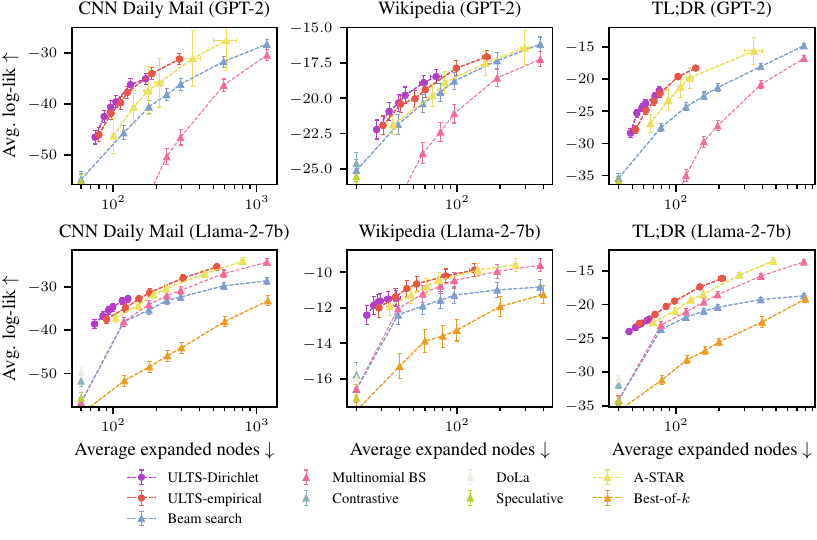}
  \caption{
    Decoding experiments with Llama-2-7b and GPT-2 for text generation on CNN Daily Mail and Wikipedia articles and text summarization for the TL;DR dataset. The methods are evaluated for different computational budgets, i.e. different values of $k$ and $k_\text{max}$. ULTS dominates the baselines in low-budget settings. Errorbars indicate two times the standard error of the mean. 
  }
  \label{fig:combined_eff}
\end{figure}

\textbf{Results} \quad Figure \ref{fig:combined_eff} shows the results. Baselines that are underperforming are shown instead in \cref{fig:nocutoff} in the Appendix.
No matter the choice of $k_{\max}$ (and \(k\)), \methodname{} yields sequences with the same or a higher log-likelihood while expanding fewer nodes. This is the case for both choices of priors, with the empirical prior encouraging exploration more.
The histogram in \cref{fig:dirichletsamples} (\cref{apd:smalleralpha}) suggests that the distribution $\max_{i} c_{ji}$ is bimodal and not fully captured by the Dirichlet prior.
In particular, our choice of Dirichlet prior is slightly too pessimistic.
As a result, the search may stop too early and the available budget may not be fully utilized.
However, the budget that \textit{is} used, is used efficiently. 
Our recommendation is thus as follows.
When efficiency is the main goal, a Dirichlet prior with low concentration is preferable---it performs similarly to beam search with smaller beam widths, while being more efficient.
If the search performance is important and an additional tree exploration can be afforded (still more efficient than beam search), then the empirical prior is the best choice---it is also hyperparameter-free. 
Despite expanding fewer nodes than beam search, ULTS is currently slower in settings where different nodes expansion in beam search can be batched. However, note that batching is not always possible, e.g. in memory-constrained settings (the memory resources depend on the model size, sequence length, as well as batch size). For more details on the runtime see \cref{apd:runtime}.

\section{Conclusion}

We have discussed \methodname{}, a probabilistic decision-making algorithm for efficiently finding high-likelihood paths on large, expensive search trees, such as those induced by LLMs. Our method quantifies and leverages computational uncertainty over node values and exploits the structure of log-likelihood rewards by placing a prior over them, using posterior samples to guide non-myopic exploration-exploitation. \methodname{} combines advantages of heuristic methods like beam search, Monte Carlo tree search, and Bayesian tree search, while crucially supporting backtracking without requiring rollouts, additional models, or complex inference. It efficiently finds high-likelihood sequences while expanding few nodes and maintaining low latency.

\textbf{Future work} \quad
One can explore batched acquisition strategies, similar to batch Bayesian optimization \citep{gonzalez2016batch, wu2016parallel}, or incorporate uncertainty over rewards to account for miscalibration. Finally, our connection between probabilistic inference and LLMs paves the way for probabilistic reasoning in the tree of thoughts framework \citep{yao2024tree}.

\section*{Acknowledgments and Disclosure of Funding}
The authors thank the International Max Planck Research School for Intelligent Systems (IMPRS-IS) for supporting JG. JG thanks Microsoft Research for support through its PhD Scholarship
Programme. PH and JG gratefully acknowledge financial support by the DFG Cluster of Excellence “Machine Learning - New Perspectives for Science”, EXC 2064/1, project number 390727645; the German Federal Ministry of Education and Research (BMBF) through the Tübingen AI Center (FKZ: 01IS18039A); and funds from the Ministry of Science, Research and Arts of the State
of Baden-Württemberg. PH gratefully acknowledges financial support by the European Research Council through ERC CoG Action 101123955 ANUBIS.
Resources used in this work were provided by the Province of Ontario,
the Government of Canada through CIFAR, companies sponsoring the Vector Institute https://vectorinstitute.ai/partners/ and the Natural Sciences and Engineering Council of
Canada. AR thanks Apple for support through the Waterloo Apple PhD Fellowship, Natural Sciences and Engineering Council of Canada for its support through the PGS-D program, and the David R. Cheriton Graduate Scholarship.

\bibliography{neurips_2025.bib}
\bibliographystyle{plain}


\appendix

\section{Further Discussions}
\label{apd:discussions}
\subsection{Prior assumptions}
The symmetry assumption for the Dirichlet prior on the LLM's softmax outputs likely does not entirely hold---some words occur more often in natural language than others \citep{zipf1932selected}.
However, \methodname{} only requires access to the distribution over the \textit{maximum} of the unexplored part of the search space and not over the \textit{arg max}.
The former is not affected by permutations of the entries in the categorical distributions, which is why we suspect that a symmetric Dirichlet distribution with sufficiently small concentration parameter is a good proxy.
Moreover, the empirical prior can also be used to address this limitation, without incurring large overhead.

We assume the same prior for all sentences in a dataset. It could be, though, that some input prompts are easier to complete than others, and the corresponding LLM's outputs are therefore generally have less entropy than on other those of other prompts.
This could be counteracted by choosing a personalized hyper-parameter $\alpha$.
This would require deriving the prior for multiple possible values of $\alpha$, which scales linearly with the number of possible values for \(\alpha\).
However, this can be precomputed (as in Alg.~\ref{alg:beta}) such that it would not affect the costs during inference.

\section{Additional Experimental Details}
\label{apd:add_details}

URLs to the models and datasets used are provided below:
\begin{itemize}
  \item Models:
        \begin{itemize}
          \item \url{https://huggingface.co/meta-llama/Llama-2-7b-hf}
          \item \url{https://huggingface.co/openai-community/gpt2}
          \item \url{https://huggingface.co/google-t5/t5-large}
          \item \url{https://huggingface.co/deepseek-ai/deepseek-coder-1.3b-base}
        \end{itemize}
  \item Datasets:
        \begin{itemize}
          \item \url{https://huggingface.co/datasets/wikipedia}
          \item \url{https://huggingface.co/datasets/cnn_dailymail}
          \item \url{https://huggingface.co/datasets/CarperAI/openai_summarize_tldr}
          \item \url{https://huggingface.co/facebook/wmt19-de-en}
          \item \url{https://huggingface.co/datasets/openai/openai_humaneval}
        \end{itemize}
\end{itemize}

\subsection{Hyperparameters}
For beam search, multinomial beam search there are no hyperparameters beyond the beam size $k$.
Speculative Search uses greedy search with n-gram based assisted decoding with \lstinline{prompt_lookup_num_tokens = 10}.
Best-of-k Sampling, we use \text{top}-p=0.95. For Nucleus Sampling, we use a temperature parameter of 0.2 and \text{top}-p=0.95. For contrastive search, we use penalty parameter $\alpha=0.6$ and \text{top}-k=50. For DoLA, we set the number of DoLA layers to "high". We use version 4.38.2. of the \texttt{transformers} package. 

\subsection{Prior Choice for ULTS}
\label{apd:prior}
We sampled a set $\mathcal{C}$ of Categorical distributions from the LLM on training sequences for each of the datasets and both of the LLMs.
We used \(1000\) samples from the training set for each of the datasets and LLMs we considered.
They are used for the empirical prior, as well as training data for fitting the concentration parameter $\alpha$ of the Dirichlet prior.

\begin{figure}
  \includegraphics[]{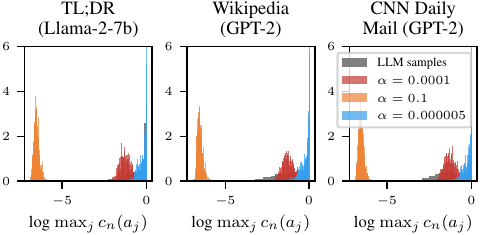}
  \caption{
    Distribution of the maximum of categorical distributions sampled from an LLM, as well as from Dirichlet priors with different concentration parameters.
  }
  \label{fig:prior}
\end{figure}

\Cref{fig:prior} shows samples for the maximum of Categorical distributions returned by the LLM, as well as the distribution of the maximum of the Categorical distribution from a Dirichlet prior for $\alpha = 10^{-1}, 10^{-4}, 5 \times 10^{-6}$.

For the experiments below, we picked a value for $\alpha$ based on visual overlap of those histogramms. For a comparison between different values of the concentration parameter, please refer to \cref{fig:smalleralpha} (\cref{apd:smalleralpha}).
Note that this way of choosing the hyperparameter $\alpha$ is very convenient compared to performing costly cross-validation.

\section{Additional Experimental Results}
\label{apd:add_results}

\subsection{On-model experiments}
\label{apd:on_model}

\begin{figure}[h!]
  \includegraphics[width=\textwidth]{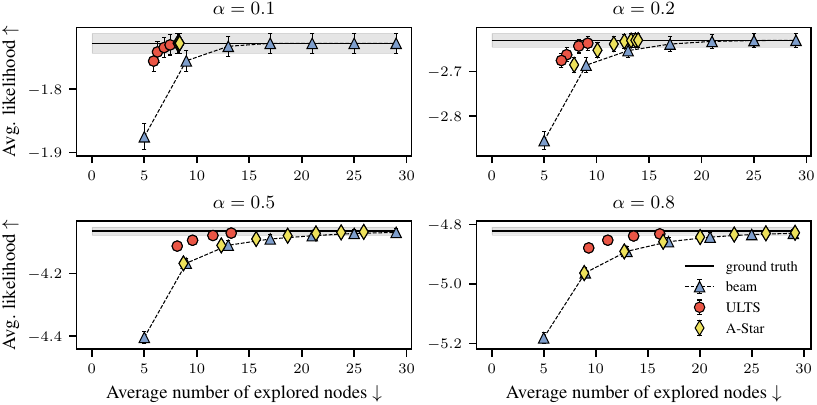}

  \caption{
    Comparison on trees, where the transition probabilities are sampled from a Dirichlet prior for different values of the concentration parameter.
  }
  \label{fig:dirichletsamples}
\end{figure}

We compare \methodname{} to beam search on artificially generated search problems from Dirichlet priors.
The trees have branching factor $B=8$ and depth $D=5$. Since these trees are so small we optimized the acquisition function in eq. (\ref{eq:acqf}) over the full boundary, i.e.\ non-recursively.
The transition probabilities at each node are sampled from a Dirichlet prior with fixed $\alpha \in \{0.1, 0.2, 0.5, 0.8\}$.
The comparison is on-model, i.e., \methodname{} is run with the ground truth parameter of $\alpha$.
We repeat the experiment with different values for the confidence parameter \(\varepsilon\) of \methodname{} from $\{0.05, 0.1, 0.3, 0.5\}$.
Since the toy problems are so small, the exploration of too many nodes is not an issue and we use $k_{\max} = \infty$.
Beam search is run with beam sizes ranging from \(1\) to \(7\).
The results in \cref{fig:dirichletsamples} show that \methodname{} dominates across the entire range of hyper-parameters.
This suggests that knowledge about concentration strength helps reduce the number of search steps.

\subsection{ULTS for code completion task}
\label{apd:codecompletion}
In addition, we test ULTS for code completion using the Deepseek Coder 1.3B LLM \citep{guo2024deepseek} and all $164$ sequences from the OpenAI HumanEval dataset \citep{chen2021evaluating}. The maximum tree depth is set to $500$ tokens and we stop at the $\langle \text{EOS} \rangle$ token. 
We use a Dirichlet prior with $\alpha =\num{5e-6}$. 
To allow for a task-specific qualitative comparison, we evaluated the generated token sequences w.r.t.\ the Pass@1 metric (i.e. percentage of test cases passed), see \cref{tab:humaneval}. 
ULTS achieves favorable performance in terms of Pass@1 and log-likelihood while expanding fewer nodes and incurring lower latency.

\begin{table*}[t]
  \footnotesize
  \centering

  \begin{tabular}{lcccc}
    \toprule
    \textbf{Method}             & \textbf{Pass@1} (\%) \(\uparrow\) & \textbf{Log-likelihood} \(\uparrow\)& \textbf{Node expansions} \(\downarrow\) & \textbf{Runtime (s)} \(\downarrow\)  \\
    \midrule
    Greedy                     & 14.02                      & -28.000          &399.369       & $13.557\pm0.493$         \\
    Nucleus sampling           & 14.63                      & -28.500        &377.671       & $10.568\pm0.429$         \\
    Beamsearch-Mult ($k=2$)    & 27.44                      & -24.375      &496.963       & $7.235\pm0.420$        \\
    Beamsearch ($k=2$)    & 29.27                     & -22.092    & 497.390      & $8.507\pm0.520$      \\
    A-STAR ($k_{max}=2)$  &      34.75                     &-16.968 & 267.841&$7.184\pm0.665$\\
    \midrule
    ULTS ($k_\text{max}=2$)               & 25.00               & -16.625        &134.835       & $4.447\pm0.421$        \\
    ULTS ($k_\text{max}=3$)               & 34.76               & -15.500        &146.866         & $4.936\pm0.454$        \\
    \bottomrule
  \end{tabular}

  \caption{
    Results for the code completion task on HumanEval.
    ULTS can achieve high performance with minimal latency.
  }
  \label{tab:humaneval}
\end{table*}

\subsection{Machine translation}
\label{apd:machinetranslation}

\begin{figure}
  \centering
  \begin{minipage}[t]{.475\textwidth}
    \centering
    \includegraphics[]{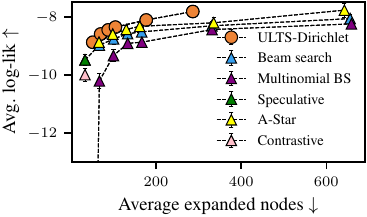}

    \vspace{-0.5em}

  \end{minipage}%
  \hfill
  \begin{minipage}[t]{.475\textwidth}
    \centering
    \includegraphics[]{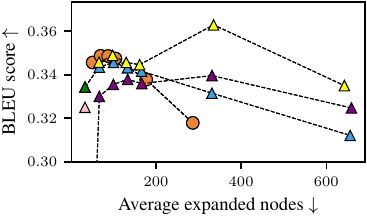}

    \vspace{-0.5em}
    \label{fig:bleu}
  \end{minipage}

  \caption{
    \textbf{Left:} Machine translation results with the WMT-19 English-to-German dataset in term of log-likelihood.  \textbf{Right:} The corresponding BLEU scores. 
  }
  \label{fig:wmt19_results_t5}
  \vspace{-1em}
\end{figure}

We compare ULTS with the baselines in a machine translation task with 1000 randomly sampled sequences from the WMT-19 English to German dataset in Figure \ref{fig:wmt19_results_t5} in terms of likelihood and BLEU score.
For the LLM, we use T5-large \citep{raffel2020t5}; for ULTS, we use a Dirichlet prior with \(\alpha = 5 \times 10^6\).
We stop the exploration of a path in the tree if the <EOS> token is found or when the maximum depth of 60 is exceeded. The results are in Figure 3. ULTS is both more efficient (fewer node expansion) and more performant (higher average log-likelihood) at all values of \(k\)/\(k_\text{max}\).
For all methods, small to medium beam sizes of $k=3$ seem to work best in terms of BLEU scores---these scores decrease for a larger (maximum) beam with, in contrast to the log-likelihood in \cref{fig:wmt19_results_t5} .

Below, we show some example sequences found with ULTS and beam search.
For these examples, we picked sentences where ULTS achieves better BLEU scores.
Other examples have mainly the same outputs as beam search while ULTS is more efficient.

\begin{mybox}{Example 1}
  \textbf{input prompt:}  \texttt{It is annoying when geographical maps are not up-to-date.}\\
  \textbf{ground truth:}  \texttt{Es nervt, wenn Landkarten nicht aktuell sind.}
  \vspace{0.3cm}
  \newline
  \textbf{ULTS translation:}  \texttt{Es ist ärgerlich, wenn geographische Karten nicht aktuell sind.}\\
  \textbf{ULTS BLEU/log-likelihood:} 0.189/-3.914 \\
  \vspace{0.3cm}
  \newline
  \textbf{Beam search translation:} \texttt{Es ist ärgerlich, wenn die geographischen Karten nicht auf dem neuesten Stand sind.}\\
  \textbf{Beam search BLEU/log-likelihood:}  0.0/-4.230 \\
\end{mybox}

\begin{mybox}{Example 2}
  \textbf{input prompt:}  \texttt{The historical maps of the digital BayernAtlas, an offering from the State Government's Geoportal Bayern, are anything but up-to-date – and yet it is precisely for this reason that they are so informative.}\\
  \textbf{ground truth:}  \texttt{Die historischen Landkarten des digitalen Bayern-Atlases, ein Angebot des Geoportals Bayern der Staatsregierung, sind alles andere als aktuell - doch gerade deshalb sehr aufschlussreich.}
  \vspace{0.3cm}
  \newline
  \textbf{ULTS translation:}  \texttt{Die historischen Karten des digitalen BayernAtlas, ein Angebot des Landesgeoportals Bayern, sind alles andere als aktuell – und gerade deshalb so informativ.}\\
  \textbf{ULTS BLEU/log-likelihood:} 0.292/-9.247 \\
  \vspace{0.3cm}
  \newline
  \textbf{Beam search translation:} \texttt{Die historischen Karten des digitalen BayernAtlas, ein Angebot des Landesgeoportals Bayern, sind alles andere als aktuell – und gerade deshalb sind sie so informativ.}\\
  \textbf{Beam search BLEU/log-likelihood:}  0.272/-9.59 \\
\end{mybox}

\begin{mybox}{Example 3}
  \textbf{input prompt:}  \texttt{Even if the French troops finally retreated with the Treaty of Lunéville from 9th February 1801: it was the current neighbours who had the idea to create a comprehensive map of Bavaria.}\\
  \textbf{ground truth:}  \texttt{Auch wenn die französischen Truppen mit dem Frieden von Lunéville vom 9. Februar 1801 schließlich abzogen: Es waren die heutigen Nachbarn, die die Idee einer flächendeckenden Bayern-Karte kreierten.}
  \vspace{0.3cm}
  \newline
  \textbf{ULTS translation:}  \texttt{Auch wenn sich die französischen Truppen mit dem Vertrag von Lunéville vom 9. Februar 1801 schließlich zurückzogen: Es waren die heutigen Nachbarn, die die Idee hatten, eine umfassende Karte von Bayern zu erstellen.}\\
  \textbf{ULTS BLEU/log-likelihood:} 0.458/-13.111 \\
  \vspace{0.3cm}
  \newline
  \textbf{Beam search translation:} \texttt{Auch wenn sich die französischen Truppen schließlich mit dem Vertrag von Lunéville vom 9. Februar 1801 zurückzogen: Es waren die heutigen Nachbarn, die die Idee hatten, eine umfassende Karte von Bayern zu erstellen.}\\
  \textbf{Beam search BLEU/log-likelihood:}  0.400/-12.754 \\
\end{mybox}

\subsection{Different choice of hyperparameter $\alpha$ for the Dirichlet prior}
\label{apd:smalleralpha}

\begin{figure}
  \begin{center}
    \includegraphics[]{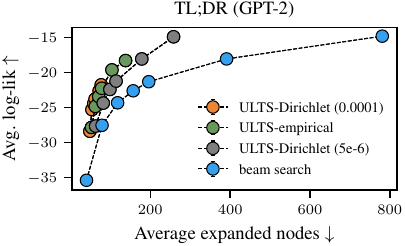}
    \caption{
      Comparison of two different values $\alpha=0.0001$ and $\alpha=5e-6$ of the Dirichlet prior conceAcquisitntration parameter in ULTS
    }
    \label{fig:smalleralpha}
  \end{center}
\end{figure}

For the Dirichlet prior, indeed $\alpha$ has a meaningful impact on the performance of ULTS: It can be seen as a hyperparameter that trades exploitation for exploitation (smaller value means more exploration). So, the choice of $\alpha$ should be based on how much budget one has since more exploration means more node expansions. Figure \ref{fig:smalleralpha} shows a comparison of two different values $\alpha=0.0001$ and $\alpha=5e-6$.

\FloatBarrier
\subsection{Number of samples in ULTS}

\begin{wraptable}[7]{c}{0.62\linewidth}
\vspace{-2em}
  \footnotesize
  \centering

  \begin{tabular}{lccr}
    \toprule
    $\bm{N}$ & \textbf{Avg. log-lik} & \textbf{node expansions} & \textbf{Time in s} \\
    \midrule
    1    & -9.454                     &86.64                         & $3.488\pm0.187$    \\
    10  & -9.066                          &90.75                & $3.631\pm0.208$    \\
    100  & -9.097                         &92.28                    & $3.690\pm0.211$    \\
    1000  & -9.074                        &92.41                    & $3.718\pm0.213$    \\
    \bottomrule
  \end{tabular}

  \caption{
    Results for different choice of number of samples $N$.
  }
\end{wraptable}

Table 2 shows results for an ablation for hyperparameter $N$ on 100 sentences from the machine translation task with maximum beam size $k_{max}=4$. Only for $N=1$ the performance is slightly worse, indicating that a small number like $N=10$ might already be sufficient.

\subsection{Runtime}
\label{apd:runtime}

\begin{figure}[h]
  \includegraphics[]{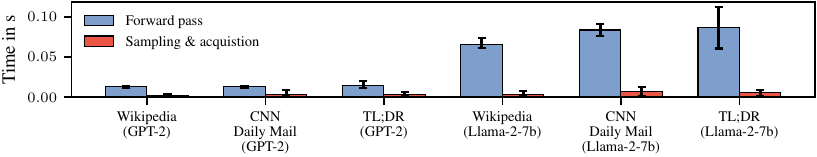}

  \vspace{-0.25em}
  \caption{
    Average wall-clock time per iteration.
    The overhead induced by ULTS is negligible.
  }
  \label{fig:runtime}
  \vspace{-0.25em}
\end{figure}

We analyze the runtime overhead on top of the LLM's forward pass due to ULTS.
\Cref{fig:runtime} shows the wall-clock time per iteration, averaged over all sentences and all iterations, broken down into the time spent on the LLM's forward pass and the time spent on sampling the optimal values and optimizing the acquisition function.
The error bars indicate the $95\%$- confidence intervals.
Results are shown for the experiments with $k_\text{max}=20$ and the Dirichlet prior---the empirical prior performs similarly since ULTS does not differentiate between them in Alg.~\ref{alg:treesearch}.
We note that the runtime overhead of ULTS is small compared to the time spent to do an LLM forward pass.
However, note that our current implementation only expands one node of the search tree in each iteration and thereby only evaluate one token sequence per forward-pass.
ULTS can be extended similar like BO can be extended into batch BO.
This is outside of the present work's scope but is a promising direction for future work.

\begin{wraptable}[7]{c}{0.5\linewidth}
  \vspace{-3em}
  \footnotesize
  \centering

  \begin{tabular}{lcr}
    \toprule
    \textbf{Method} & \(\bm{k}\) \textbf{or} \(\bm{k_\mathrm{max}}\) & \textbf{Time in s} \\
    \midrule
    Beam search     & 5                                              & $2.158\pm0.029$    \\
    ULTS Dirichlet  & 3                                              & $4.622\pm0.114$    \\
    ULTS Dirichlet  & 5                                              & $5.177\pm0.164$    \\
    \bottomrule
  \end{tabular}

  \caption{
    Total runtime for decoding one of the TL;DR input prompts with Llama-2-7b.
  }
\end{wraptable}

In Table 3 we show total runtime results for beam search with beam size $k=5$ and ULTS with maximal beam size $k_{max}=3$ and $k_{max}=5$. Despite expanding fewer nodes than beam search, ULTS is currently slower in settings where different nodes expansion in beam search can be batched. However, note that batching is not always possible, e.g. in memory-constrained settings (the memory resources depend on the model size, sequence length, as well as batch size).\\

As a reference regarding the computation of the prior:
Building the Dirichlet Prior for a tree of depth 250 an branching size 32256 with 1000 samples on a desktop machine (MacBook M1) with CPU only takes 10:50 min (2.6 secs per level of the tree). 

\subsection{Alternative acquisition function}
\label{apd:posterior_acqf}

As discussed in \cref{acquisition function}, different selection (and hence backup) strategies can be utilized.
Recall that all of our results so far are obtained using the ``posterior descendant'' strategy defined in \eqref{eq:acqf_recurrence}.
Here, we show the corresponding results where the other strategy (``posterior'') is used instead.

\begin{figure}
  \includegraphics[width=\textwidth]{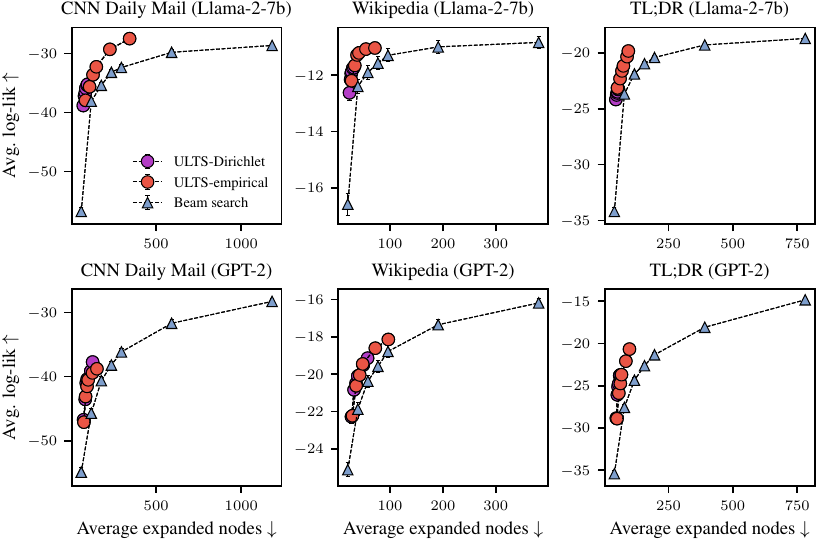}
  \caption{
    Decoding experiments with Llama-2-7b and GPT-2 with the ``posterior'' strategy.
  }
  \label{fig:combined_eff_posterior}
\end{figure}

First, \cref{fig:combined_eff_posterior} shows results under the same setting as in the main text, but both ULTS-Dirichlet and ULTS-Emp use the ``posterior'' strategy instead.
We noticed that this strategy also performs well---it is more efficient than beam search.
Moreover, it also achieves better or similar likelihood than beam search in Llama-2-7b.

We further compare the ``posterior'' strategy compared to the ``posterior descendant'' strategy in \cref{fig:headtohead}.
We notice that the ``posterior'' strategy tends to under-explore compared to the ``posterior descendant'' strategy. Hence, we use and recommend the ``posterior descendant'' strategy by default.
In general, the optimal value estimates from the ``posterior descendant'' strategy are less optimistic due to Jensen inequality. This effect is larger if the optimal values of children lie closer together. We illustrate this with the following two examples: In the first example, one of the children is much better than the other children. In such scenarios there is no big difference between the back-up-ed values. In the second example, where some children are close together in terms over their optimal value estimates, the ``posterior descendant'' estimate is lower making it more likely for the other siblings to get explored, too.  We therefore hypothesize that the overall effect of the ``posterior descendant'' strategy is that exploration is a bit more encouraged as also observed empirically in Figure \ref{fig:headtohead}. 

\begin{figure}
  \centering
  \begin{minipage}[t]{.475\textwidth}
    \centering
    \includegraphics[width=\textwidth]{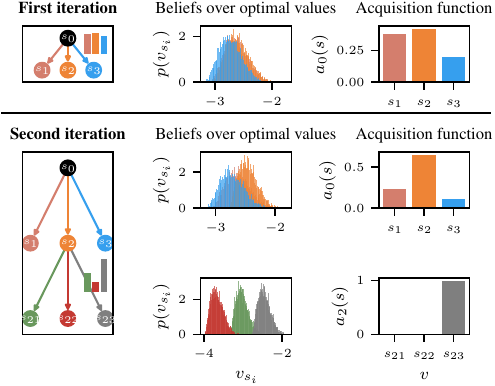}

    \vspace{-0.5em}

  \end{minipage}%
  \hfill
  \begin{minipage}[t]{.475\textwidth}
    \centering
    \includegraphics[width=\textwidth]{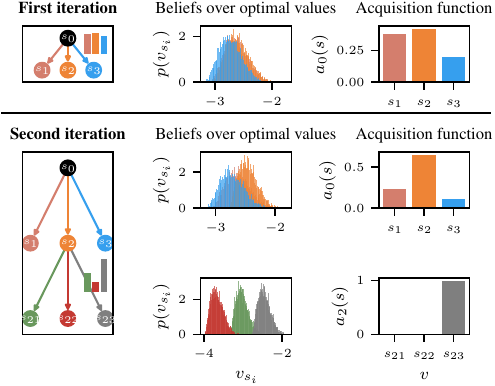}

    \vspace{-0.5em}
    \label{fig:bleu}
  \end{minipage}

  \caption{\textbf{Example 1:}
    \textbf{Left:} ``posterior descendant'' acquisition function.\textbf{Right:} ``posterior'' acquisition function.
  }
  \label{fig:wmt19_results_t5}
  \vspace{-1em}
\end{figure}

\begin{figure}
  \centering
  \begin{minipage}[t]{.475\textwidth}
    \centering
    \includegraphics[width=\textwidth]{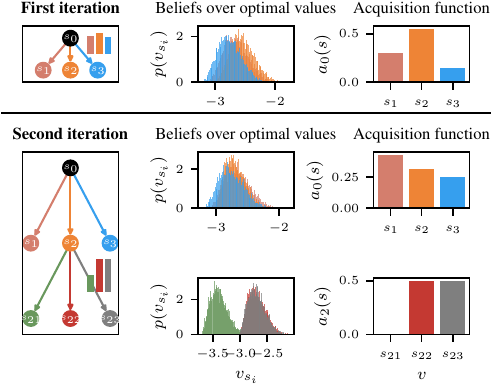}

    \vspace{-0.5em}

  \end{minipage}%
  \hfill
  \begin{minipage}[t]{.475\textwidth}
    \centering
    \includegraphics[width=\textwidth]{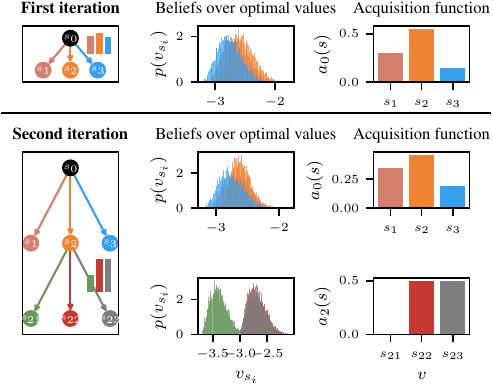}

    \vspace{-0.5em}
    \label{fig:bleu}
  \end{minipage}

  \caption{\textbf{Example 2:}
    \textbf{Left:} ``posterior descendant'' acquisition function \textbf{Right:} ``posterior'' acquisition function.
  }
  \label{fig:wmt19_results_t5}
  \vspace{-1em}
\end{figure}

\begin{figure}
  \includegraphics[width=\textwidth]{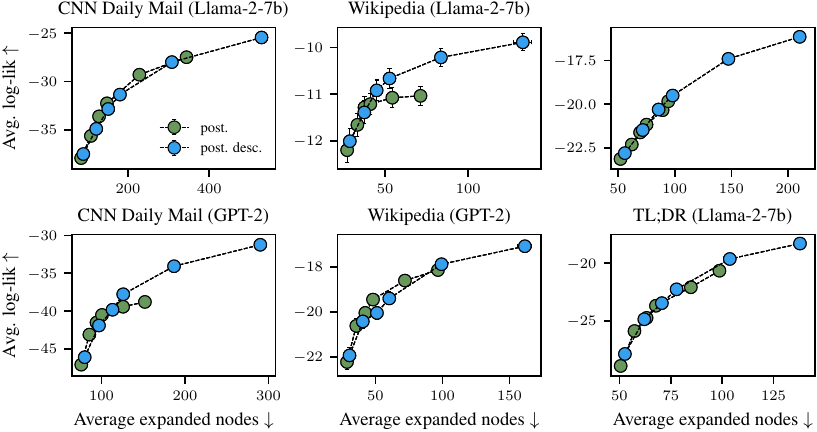}
  \caption{
    ``Posterior'' vs.\ ``posterior descendant'' acquisition function.
  }
  \label{fig:headtohead}
\end{figure}

\subsection{Full results for decoding experiments with LLama-2-7b and GPT-2}
\label{apd:nocutoff}

\begin{figure}
  \includegraphics[width=\textwidth]{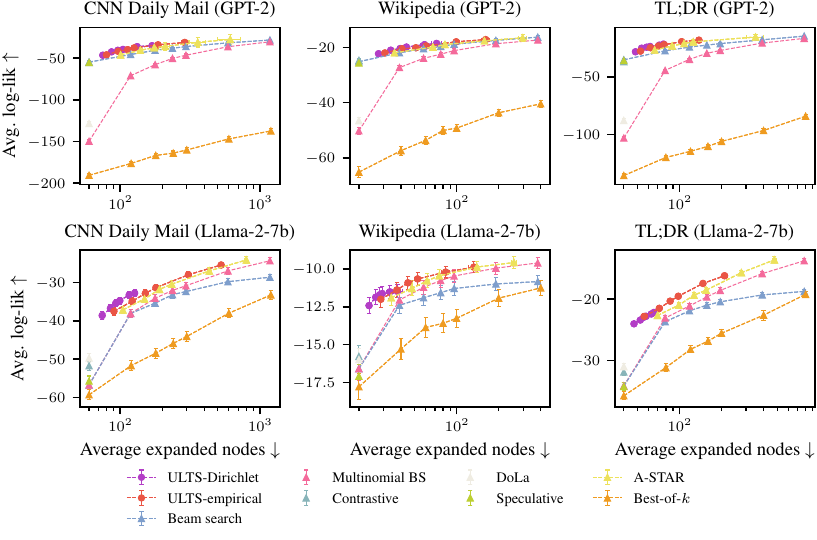}
  \caption{
    Full results for decoding experiments with LLama-2-7b and GPT-2
  }
  \label{fig:nocutoff}
\end{figure}

Figure \ref{fig:nocutoff} shows the full results from the decoding experiments with LLama-2-7b and GPT-2 from section \ref{sec:experiments} in the main text.

\subsection{Experiment with alternative utility function}
\label{apd:penalty}
In the following experiment, we evaluate ULTS with a different utility function -- one that additionally contains a repetition-penalty term:
\begin{align}
\label{eq:diversityutility}
\tilde{u}(x_c, \vv_i) = \mathbb{I} [ (\log v_{\hat{x}_c} + \lambda \, b(\hat{x}_c)) = \max_{x_j \in \mathrm{children}(x_i)} (\log v_{\hat{x}_j} + \lambda \, b(\hat{x}_j)) ],
\end{align}
where \(b(\hat{x}_c)\) is a log-diversity term \citep{su2022contrastiveframeworkneuraltext}.
In this task, we complete 100 sentences from the Wikipedia dataset with GPT-2. We predict 100 tokens. We define the diversity term $b(x)$ of a node $x$ based on the proportion of duplicated $n$-grams in the token sequence $(a_1, ..., a_i)$ corresponding to the node. For a token sequence $(a_1, ..., a_i)$, let $\textbf{rep-n}((a_1, ..., a_i)) = \bigl(1- \frac{|\text{unique $n$-grams}((a_1, ..., a_i))|}{|\text{total $n$-grams}((a_1, ..., a_i))|}\bigr)$. Following \cite{su2022contrastiveframeworkneuraltext}, we measure diversity by taking repetitions at different $n$-gram levels into account:
\begin{align*}
  \textbf{diversity}((a_1, ..., a_i)) = \prod_{n=2}^4 (1- \textbf{rep-n}((a_1,...,a_i))).
\end{align*}
Since we ULTS optimizes the total likelihood of a sequence and not the average likelihood of a sequence, we additionally scale the term with the tree depth, leading to $b(x) = d \cdot \log \textbf{diversity}((a_1, ..., a_i))$ in log-space.
For ULTS we use a Dirichlet prior with concentration parameter $\alpha=5\times10^{-6}$, $\epsilon=0.1$, $k_{max}=5$. and penalty parameter $\lambda \in \{0.0, 0.2, 0.4, 0.6, 0.8, 1.0\}$. The parameters for contrastive search are top-$k$ $= 5$ and penalty parameter $\alpha = 0.6$. For beam search, we use $k=5$. The main paper reports results for ULTS with $\lambda = 0.0$, i.w. without penalty, and $\lambda= 0.2$. The full results are given in Table 3. One can see that with increasing penalty the diversity of the decoded sequences increases, i.e. extending the acquisition function with a repetition term is effective.

\begin{table}
  \footnotesize
  \centering

  \begin{tabular}{lrrr}
    \toprule
    \textbf{Method}             & \textbf{Perplexity} \(\downarrow\) & \textbf{Diversity} \(\uparrow\) \\
    \midrule
    Beam search                 & 1.16$\pm$0.01                      & 0.33$\pm$0.01                   \\
    Nucleus sampling            & 3.69$\pm$0.07                      & 0.82$\pm$0.01                   \\
    Contrastive search          & 1.27$\pm$0.01                      & 0.4$\pm$0.01                    \\
    ULTS                        & 1.16$\pm$0.01                      & 0.33$\pm$0.01                   \\
    ULTS with \(\lambda = 0.2\) & 1.3$\pm$0.01                       & 0.55$\pm$0.01                   \\
    ULTS with \(\lambda = 0.4\) & 1.45$\pm$0.01                      & 0.74$\pm$0.01                   \\
    ULTS with \(\lambda = 0.6\) & 1.5$\pm$0.01                       & 0.79$\pm$0.01                   \\
    ULTS with \(\lambda = 0.8\) & 1.53$\pm$0.01                      & 0.82$\pm$0.01                   \\
    ULTS with \(\lambda = 1.0\) & 1.55$\pm$0.01                      & 0.83$\pm$0.01                   \\
    \bottomrule
  \end{tabular}

  \caption{
    Effects of a penalty term in ULTS' acquisition function.
  }
  \label{tab:penalty_full}
\end{table}

\subsection{Additional metrics}
\label{apd:rouge}

\begin{figure}
\includegraphics[]{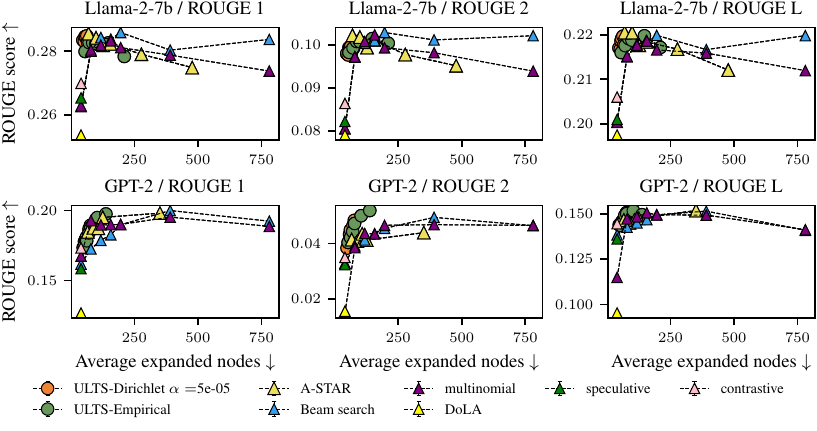}
\caption{ROUGE scores for the summarization task.}
\label{fig:rouge}
\end{figure}
\begin{figure}
\begin{center}
\includegraphics[]{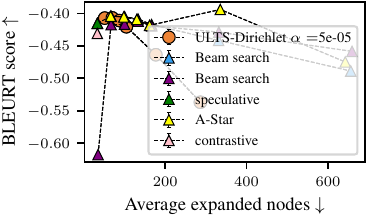}
\caption{BLEURT scores for the machine translation task.}
\label{fig:bleurt}
\end{center}
\end{figure}

Figure \ref{fig:rouge} shows an evaluation in terms of ROUGE metrics for the summarization task on the TL;DR dataset described in Section \ref{sec:experiments}. ROUGE 1 quantifies the overlap of unigrams between the produced sentence and ground truth reference, ROUGE 2 refers to bigrams, and ROUGE L refers to the longest commen subsequence. Though there is no perfect correlation between the ROUGE metric and the log-likelihood, ULTS still tends to achieve a good trade-off between performance and node expansions overall.
Figure \ref{fig:bleurt} shows additional results in terms of the BLEURT metric. Here, ULTS performs well for small $k_{max}$ but shows a decrease in performance for larger $k_{max}$. 

\subsection{ULTS with larger budget}
\label{apd:largerbudget}
\begin{figure}
\centering
\includegraphics[]{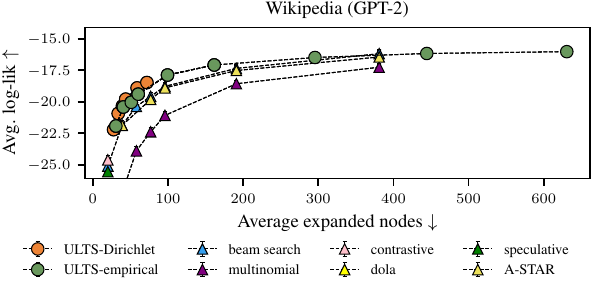}
\caption{Comparison of ULTS with the baselines on the Wikipedia dataset as described in Section \ref{sec:experiments} with additional data points for $k_{\max} \in \{50, 100, 200\}$.}
\label{fig:wiki_large}
\end{figure}

\begin{figure}
\centering
\includegraphics[]{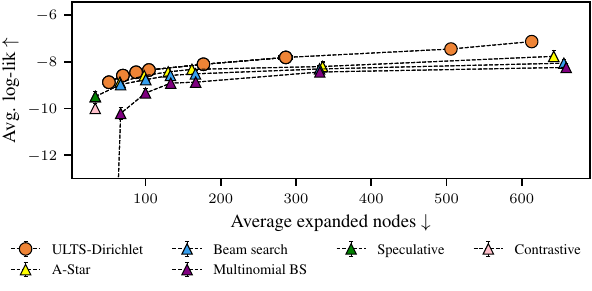}
\caption{Comparison of ULTS in the machine translation task as described in Section \ref{sec:experiments} with additional data points for $k_{\max} \in \{50, 100\}$.}
\label{fig:wmt_large}
\end{figure}

Figure \ref{fig:wiki_large} shows additional results for ULTS with the empirical prior and GPT-2 on the Wikipedia dataset, as well as ULTS with the Dirichlet prior and t5 on the machine translation task (Figure \ref{fig:wmt_large}).

\section{Pseudocode}
\label{apd:pseudocode}

\begin{algorithm}
    \caption{\methodname{}}
    \label{alg:treesearch}
    \begin{algorithmic}[1]
        \REQUIRE Number of tokens to generate $D$, number of samples $N$, approximate priors $\{\mathcal{B}_l\}_{l=1}^D$, confidence parameter $\varepsilon$, maximum number of expandable nodes per level $k_{\text{max}}$
        \ENSURE A leaf $s^*$ and its likelihood $c^*$
        
        \COMMENT{Initialization}
        \STATE $\mathcal{L} \gets \{s_0\}$
        
        \FOR{$n \gets 1$ to $N$}
            \STATE Sample $\Delta_{s_0, n} \sim \mathcal{B}_{1}(\Delta)$
            \STATE $v_{s_0, n} \gets \log \Delta_{s_0, n}$
        \ENDFOR
        \STATE $c^* \gets -\infty$, $s^* \gets \text{None}$

        \WHILE{$\hat{\mathbb{P}}(c^* < v_{s_0}) > \varepsilon$}
            \STATE \textcolor{mycolor}{\texttt{//Selection always starts from root}}\\
            \STATE $s_i \gets \texttt{select}(s_0, k_{\text{max}})$\\
            
            \textcolor{mycolor}{\texttt{//Expand}}\\
            \STATE $\mathcal{L} \gets (\mathcal{L} \setminus \{s_i\}) \cup \texttt{children}(s_i)$\\
            
            \FOR{$s_c \in \texttt{children}(s_i)$}
                \STATE \textcolor{mycolor}{\texttt{//Generate posterior samples}}
                \FOR{$n \gets 1$ to $N$}
                    \STATE Sample $\Delta_{s_c, n} \sim \mathcal{B}_{\texttt{level}(s_c)}(\Delta)$
                    \STATE $v_{s_c, n} \gets \log c_{s_0 \to s_c} + \log \Delta_{s_c, n}$
                \ENDFOR

                \STATE \textcolor{mycolor}{\texttt{//Update best complete path so far}}
                \IF{$\texttt{level}(s_c) = D$ \AND $\log c_{s_0 \to s_c} > c^*$}
                    \STATE $c^* \gets \log c_{s_0 \to s_c}$
                    \STATE $s^* \gets s_c$
                \ENDIF
            \ENDFOR

            \STATE $\texttt{backup}(\{v_{s_c, n}\}_{n=1}^{N}, s_0 \to s_c)$

            \STATE \textcolor{mycolor}{\texttt{//Termination probability}}
            \STATE $\hat{\mathbb{P}}(c^* < v_{s_0}) \gets \frac{1}{N} \sum_{n=1}^{N} \mathbb{I}[c^* \leq v_{s_0, n}]$
        \ENDWHILE
    \end{algorithmic}
\end{algorithm}


\end{document}